\def\year{2015}
\documentclass[letterpaper]{article}
\usepackage{aaai}
\usepackage{amsmath,amsthm,amssymb,amsfonts}
\usepackage{graphicx}
\usepackage{algorithm} 
\usepackage{algpseudocode}
\usepackage{subcaption}
\usepackage{url}
\usepackage[colorlinks=true]{hyperref}
\usepackage{times}
\usepackage{helvet}
\usepackage{courier}
\newtheorem{definition}{Definition}

\begin{document}
%
\title{Stochastic Blockmodeling for Online Advertising}
\author{Li Chen \\
Department of Applied Mathematics and Statistics \\ Johns Hopkins University\\ Baltimore, MD 21210 \\lichen.jhu1@gmail.com
\thanks{This work was performed when the first author was a summer intern at AOL.}
\And
Matthew Patton \\ AOL Advertising.com \\ 1020 Hull St \\ Baltimore, MD 21230 \\ matthew.patton@teamaol.com}

\maketitle

\section{Introduction}
Online advertising is market communication over the internet. This form of advertising has proven its importance in the golden digital age. 
In this work, we approach the business problem of improving the effectiveness of ad campaigns and targeting a wider range of online audience via discovering the intrinsic structures of the websites. 
We introduce a stochastic blockmodeling framework for discovering the website structures, propose two vertex clustering algorithms based on the Bayesian information criterion, and compare the performance with a goodness-of-fit method and a deterministic graph partitioning method. We demonstrate the effectiveness of our proposed algorithms on simulation and the AOL website dataset. 

\section{Background}
A graph $G = (V, E)$ represents a collection of interacting objects, where the objects are vertices in the vertex set $V:=\{1,2,\dots,\}$  containing $n$ vertices, and the interactions are edges in the edge set $E$. The adjacency matrix $A$ of the graph $G$ is an $n \times n$ matrix. In this work, we assume that the graph $G$ is undirected, unweighted, and not loopy. That implies the adjacency matrix $A$ is symmetric, binary and hollow. A random graph is a graph-valued random variable: $\mathcal{G}: \Omega \rightarrow \mathcal{G}_n$, where $\mathcal{G}_n$ represents the collection of all $2^{{n \choose 2}}$ possible graphs on the vertex set $V$. Associated with the adjacency matrix $A $, there exists a communication probability matrix $P \in [0,1]^{n \times n}$, where each entry $P_{ij}$ denotes the probability of edge existence between vertex $i$ and vertex $j$. 

One very general random graph is the latent position graph \cite{hoff2002latent}. Each vertex $v_i$ is associated with a latent position $X_i$ drawn independently from some distribution $F$ on $\mathbb{R}^D$. The edge probability between vertices $i$ and $j$ is $P_{ij} = \text{Bernoulli}(l(X_i, X_j))$, where $l: \mathbb{R}^D \times \mathbb{R}^D \rightarrow [0,1]$ is the link function. The stochastic blockmodel \cite{holland1983stochastic} is a special case of the latent position graph, where its latent positions are point masses. The vertices are partitioned in to $K$ blocks, and the edges are conditional independent Bernoulli trials. 
\begin{definition}\textbf{Stochastic blockmodel (SBM)}
Let $K$ be the number of block memberships. Let $\pi$ be the block membership function of length $K$ vector such that $\sum_{k=1}^{K}\pi_k = 1$. Let $\{Y_i\}_{i=1}^{n} \sim \text{Categorical}([K], \pi)$ be the block memberships. Let $B \in [0,1] ^{K \times K}$ be a symmetric probability matrix denoting the block probabilities. Then $A \sim SBM([n], B, \pi)$ if 
\begin{eqnarray*}
\mathbf{P}_{ij} &=& Prob(A_{ij}= 1 |X_i, X_j) \\
& = & Prob(A_{ij} = 1 |Y_i, Y_j) = B_{Y_i, Y_j}. 
\end{eqnarray*}
\end{definition}
\section{The BIC-Based Vertex Clustering Algorithms}
In the classical setting for unsupervised learning, we observe independently identically distributed feature vectors $X_1, X_2, \dots, X_n$, where each $X_i: \Omega \rightarrow \mathbb{R}^{D}$ is a random vector for some probability space $\Omega$. Here we consider the case when the feature vectors $X_1, X_2, \dots, X_n$ are unobserved. Instead we observe a latent position random graph $\mathcal{G}(X_1, X_2, \dots, X_n)$ on $n$ vertices. We intend to cluster the vertices using the observed graph.

For stochastic blockmodels with $K$ blocks and a known model dimension $D$, \cite{sussman2012consistent} and \cite{rohe2011spectral} respectively have shown that adjacency spectral embedding and Laplacian spectral embedding are consistent estimates of the latent positions. The resulted embedding is a $K$-mixture and $D$-variate Gaussian distributions asymptotically. Such results motivate us to propose a model-based clustering approach on the embedded space of the stochastic blockmodel. The optimal number of clusters and covariance structure correspond to the model selection criterion by the Bayesian information criterion (BIC). Our approach is presented in Algorithm \ref{alg:bisca1}. We denote the algorithms by ASE and LAP respectively, if $M$ is either the adjacency matrix or the Laplacian matrix. 

\begin{algorithm}[H]
\caption{The BIC-based Vertex Clustering Approach}
\begin{algorithmic}
\State \textbf{Input:} An input square matrix $M$ of order $n$, an integer $K \geq 1$, and an embedding dimension $D$.
\State \textbf{Step 1} :  Compute the first $D$ orthonormal eigenpairs of $M$, denoted by $(U_M , S_M )  \in \mathbb{R}^{n \times D} \times \mathbb{R}^D$. 
\State \textbf{Step 2:} Define the $D$-dimensional embedding of $M$ to be $\hat{M} := U_M S_M^{1/2}$.
\State \textbf{Step 3:} \For {$k$ in $1:K$} 
\State Fit Gaussian mixture models with different covariance types and $k$ clusters to $\hat{M} $, and compute the BIC. 
\EndFor
\State \textbf{Step 4:} Cluster the vertices using the optimal model selected via the maximum BIC.
\end{algorithmic}
\label{alg:bisca1}
\end{algorithm}

We compare ASE and LAP with the integrated classification likelihood (ICL) method \cite{daudin2008mixture}, which is a likelihood maximization method for stochastic blockmodels, and the Louvain  algorithm \cite{blondel2008fast}, which optimizes graph modularity and performs efficiently for large graphs.

\section{Experiments}

\subsection{Clustering Validation} \label{subsec: clust_eval}
For simulation, we measure clustering performance using the adjusted rand index (ARI) \cite{hubert1985comparing}. The higher ARI indicates a better clustering performance. For real data experiment, one challenge for clustering validation is the lack of ground truth. We examine the significance of the clusters using external datasets including website topics, revenue per website, number of clicks on ads per website, and impressions (the volume of ad display) per website. 


\subsection{Simulation}
In this experiment, only ASE and LAP correctly select the number of blocks with an average ARI above $90\%$. As the number of clusters increases, the ARIs of ASE and LAP decrease due to the phenomenon of bias-variance tradeoff.

\subsection{AOL Website Graph}
We use the relational events of online users who visit the websites during the day July 1, 2014 under one campaign, and build a graph containing websites as vertices. An edge exists between websites $v_i$ and $v_j$, if they share at least one common user. The resulted adjacency matrix is symmetrized, binarized, hollow, and of size $1569\times 1569$. 
In practice, the true model dimension $D$ of the stochastic blockmodel is unknown. We estimate $D$ using a profile likelihood maximization method \cite{zhu2006automatic}. 


The ARIs for comparing the $6$ pairs of algorithms indicate the partitions by ASE and LAP are most similar, while ASE detects most block signal. The partitions by ICL and Louvain are least similar. In terms of website topics, Cluster $1$ discovered by ASE mainly contains references and popular sites. Cluster $5$ discovered by ASE mainly contains websites on politics, baby, teen, gallary. The clusters discovered by the other clustering algorithms do not correspond to any significant topic clusters. 

In addition, we evaluate the clusters using revenue, clicks and impressions. For each pair of clusters in each clustering algorithm, we apply two-sided Wilcoxon rank sum test with the null hypothesis that the revenue/clicks/impressions are the same. At a significance level of $5\%$, ASE discovers more statistically significant clusters based on the business metrics than other algorithms. 

See \url{https://sites.google.com/site/lichenjhuresearch/home} for details.




\section{Discussion}
In this work, we introduce a stochastic blockmodeling framework for online advertising, and propose two BIC-based vertex clustering algorithms: ASE and LAP. We demonstrate in simulation that our proposed algorithms are able to detect the correct number of blocks. In the AOL website graph experiment, our proposed algorithm ASE is able to detect significant website clusters validated via impressions, revenue and number of clicks. The applications of our approach not only extend to further cluster-based inference, but also can serve as a basic guidance for website inventory acquisition. While our proposed approach is presented for undirected and unweighted graphs, it adapts to directed and weighted graphs. We are optimistic that random graph framework is valuable for online advertising research.

\bibliographystyle{aaai}
\bibliography{shortPaper}
\end{document}